\documentclass[runningheads]{llncs}
\usepackage[T1]{fontenc}
\usepackage{mathtools} 
\DeclareMathOperator{\SDF}{SDF}
\DeclareMathOperator*{\argmin}{arg\,min}
\newcommand{\clamp}[1]{\left[#1\right]_{-\delta}^{\delta}}

\usepackage{graphicx,verbatim}
\usepackage{xcolor}
\usepackage{hyperref}
\usepackage[capitalise,nameinlink]{cleveref} 
\crefname{figure}{Fig.}{Figs.} 
\Crefname{figure}{Figure}{Figures}
\crefname{table}{Table}{Tables}
\Crefname{table}{Table}{Tables}
\usepackage{subcaption}
\usepackage{multirow}
\usepackage{booktabs}
\usepackage{microtype}

\begin{document}
\title{Causal Spatio-Temporal Neural Distance Fields for Counterfactual Cardiac Anatomy}
%
\author{%
Mathias Lowes\inst{1}\orcidID{0009-0003-0002-1645} \and
Kurt Butler\inst{2,3}\orcidID{0000-0002-1520-4909} \and
Kristine Aavild S{\o}rensen\inst{1,4}\orcidID{0000-0003-1434-697X} \and
Klaus F. Kofoed\inst{5}\orcidID{0000-0001-9742-1554} \and
Sotirios A. Tsaftaris\inst{2,3}\orcidID{0000-0002-8795-9294} \and
Rasmus R. Paulsen\inst{1}\orcidID{0000-0003-0647-3215} \and
Steven McDonagh\inst{2,3}\orcidID{0000-0001-7025-5197}%
}

\authorrunning{M. Lowes et al.}
\titlerunning{Causal Neural Distance Fields for Counterfactual Cardiac Anatomy}

\institute{%
DTU Compute, Technical University of Denmark, Kongens Lyngby, Denmark \\
\email{mmilo@dtu.dk} \and
Causality in Healthcare AI Hub (CHAI), UK \and
School of Engineering, University of Edinburgh, Edinburgh, UK \and
Novo Nordisk A/S, Copenhagen, Denmark \and
Department of Cardiology, Rigshospitalet, Copenhagen University Hospital,
Copenhagen, Denmark
}
  
\maketitle              

\begin{abstract}
Counterfactual cardiac anatomy can support risk modelling and intervention planning, by predicting how a patient's heart would change under altered clinical conditions. However, current shape generative models are conditional on clinical variables and disregard causal links between these variables. Thus, they cannot answer counterfactual questions about how a given subject's anatomy would change under a different clinical profile. 
Modelling the heart's continuous movement over the cardiac cycle requires a representation that is itself continuous in space and time. Spatio-temporal neural distance fields provide this while separating clinical conditioning from subject-specific shape.
We place a structural causal model (SCM) over the set of clinical variables (age, sex, smoking, BMI, heart rate, systolic blood pressure), before the condition encoding of the spatio-temporal neural distance field which represents the left atrial complex over a cardiac cycle by signed distance fields. 
An intervention of the clinical variables, propagated through a SCM before the 4D counterfactual, is constructed from the neural distance field. 
The generated counterfactuals follow dose-response under age interventions and transforms the male distribution to closely match the female distribution under the intervention $\mathrm{do(\mathrm{female}{=}1)}$. By doing inverse interventions the model also approximates the original anatomy with sub millimeter precision.
To our knowledge, this is the first counterfactual generation of 4D cardiac anatomy and the first integration of an SCM with an implicit neural representation as its imaging mechanism.



\end{abstract}
\section{Introduction}
The heart is a complex three-dimensional anatomical structure that continuously deforms throughout the cardiac cycle, giving rise to 3D+t (4D) anatomy. Capturing the shape and motion of complex heart chambers such as the left atrium is central to the study of cardiac function and disease. 
Voxel- and mesh-based representations scale poorly for 4D anatomy requiring either a fixed grid that scales memory or shape correspondence across subject, which varying anatomy makes infeasible.
Spatio-temporal neural distance fields have emerged as a compact, continuous representation that models 4D cardiac anatomy as as sequence of signed distance fields (SDFs), without being locked to a grid or needing point correspondence~\cite{sorensen2024stndf}. These representations further separate clinical latent code from subject specific latent codes, enabling interventions that alter clinical variables while holding individual anatomy fixed.

Comprehensive 4D analysis requires diverse data to reflect the many ways in which cardiac anatomy and motion vary across individuals. Because high-quality, annotated 4D representations of the heart are difficult to acquire, generative models have been widely adopted to alleviate this scarcity. However, standard generative models inherit biases from their training data, and do not automatically respect anatomical constraints or medical interventions. Causal generative models, in contrast, address this gap by explicitly modelling causal relationships between variables and decoupling spurious correlations \cite{castro2020causality}.

Causal generative modelling of medical images has been established on grid-based representations, mostly from the development of the deep structural causal model (DSCM) framework~\cite{desousaribeiro2023highfidelity,pawlowski2020dscm}.
Meanwhile, conditional generative models of temporal cardiac anatomy do exist both for implicit representations~\cite{sorensen2024stndf} and for grid representations~\cite{Qiao2024Cheart}. However, these do not incorporate clinical information in a causal fashion, posing the risk that spurious correlations between clinical features might be mistaken for biological mechanisms. A causal generative model for temporal cardiac data has been established for 2D+t ultrasound pixels, D'Artagnan \cite{Reynaud2022dartagnan}, which conditioned on a single scalar.

We propose an approach to causally generate spatio-temporal cardiac anatomy, by placing a structural causal model (SCM) over the clinical variables, before the clinical encoding of the spatio-temporal neural distance field. An intervention propagates through the SCM, making the clinical variables remain mutually consistent rather than altered in isolation, before conditioning the network, thus the generated 4D anatomy becomes a causally consistent counterfactual for the individual. 
We demonstrate this approach on the left atrial complex, consisting of left atrium, left atrial appendage and pulmonary veins, chosen for data availability. However, the method makes no assumptions specific to this anatomy and is in principle applicable to the whole heart and other organs, as long as they can be represented by an SDF. Our main contribution entails the first combination of an SCM with an implicit neural representation, yielding a causal conditional implicit neural network.

\section{Methods}
\begin{figure}[ht]
    \centering
    \includegraphics[width=\linewidth]{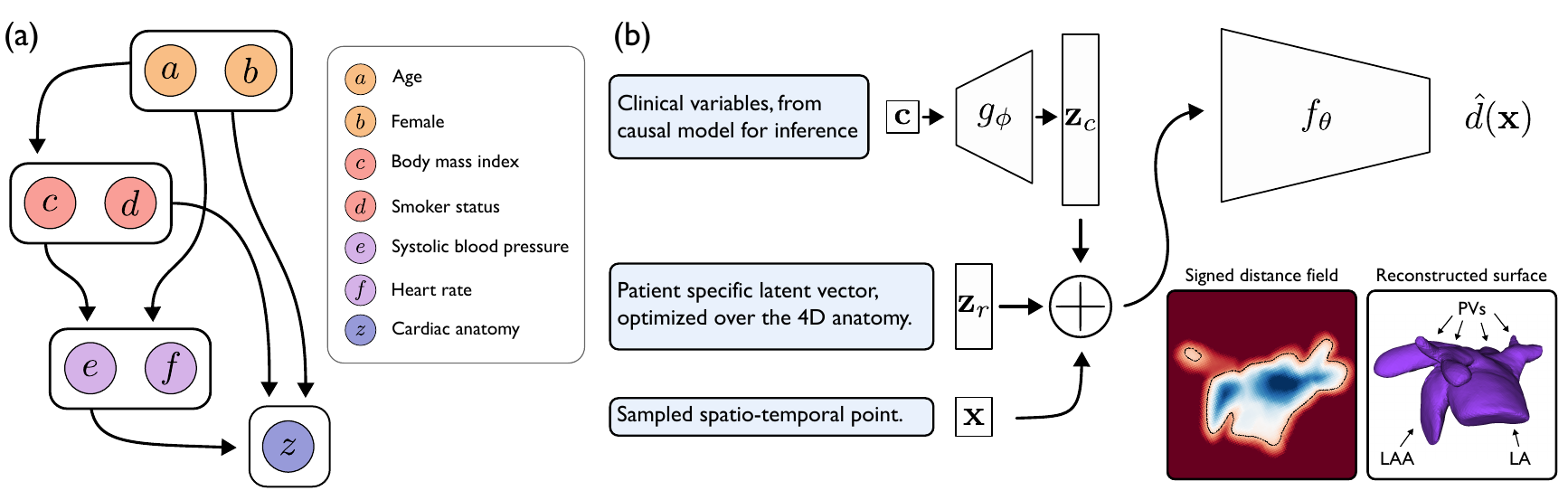}
    \caption{(a) proposed causal graph for cardiac anatomy, where the nodes are grouped to simplify the visualization. 
    An edge between two groups of nodes, $A \to B$, indicates that every node in A connects to every node in B.
    (b) clinical variables are encoded to the latent space  $\mathbf{z}_c = g_\phi (\mathbf{c})$ and concatenated with the patient latent $\mathbf{z}_r$ and spatio-temporal point $\mathbf{x}$ to estimate the signed distance  $\hat d (\mathbf{x}) = f_\theta(\mathbf{x},\mathbf{z}_c,\mathbf{z}_r)$.
    } 
    \label{fig:method}
\end{figure}
\textbf{Overview.}
We model the four-dimensional (3D + time) anatomy of the left atrial complex, consisting of the left atrium, left atrial appendage, and pulmonary veins (LA+LAA+PV), as the terminal node of an SCM defined over a set of clinical variables. The surface is represented implicitly by a neural signed distance field (SDF) which maps a spatio-temporal point $\mathbf{x}=(\mathbf{p},t)$ to signed distance as $\hat d (\mathbf{x}) = f_\theta(\mathbf{x},\mathbf{z}_c,\mathbf{z}_r)$. The function $f_\theta$ is the implicit representation of the signed distance field of the cardiac anatomy, where the  zero level set traces the anatomical surface. Each subject is described by the two latent codes $\mathbf{z}_c$ and $\mathbf{z}_r$ following the architecture of Sørensen et al.~\cite{sorensen2024stndf}. Here, $\mathbf{z}_c$ encodes information from the subjects clinical variables and $\mathbf{z}_r$ is a learned subject-specific latent code. See \cref{fig:method} for an overview of our method.

We extent the work of Sørensen et al.~\cite{sorensen2024stndf} by placing a SCM over the clinical variables before the encoding $\mathbf{z}_c = g_\phi (\mathbf{c})$. 
Without the SCM, changing a clinical variable amounts to editing a single entry of
$\mathbf{c}$ in isolation. By modelling $\mathbf{c}$ as the output of an SCM, an intervention instead propagates through the clinical mechanisms before encoding. Thus, the conditioning reflects a causally consistent counterfactual rather than an associational edit.\footnote{Code available at: \url{https://github.com/MMLowes/spatio_temporal_causal_cardiac_model}}

\textbf{Spatio-temporal Neural Distance Field.}
The model architecture is adopted from~\cite{sorensen2024stndf} with a clinical encoder $g_\phi$ and a decoder $f_\theta$. The decoder is trained as a conditional auto-decoder, which takes the spatio-temporal point $\mathbf{x}$, the clinical encoding $\mathbf{z}_c$, and a subject specific latent vector $\mathbf{z}_r$ as input. The subject latents are optimized jointly with $\theta$ under a Gaussian prior by an $L_2$ loss, see \cite{park2019deepSDF} for more information. Following recent studies on INR for periodic data~\cite{Lowes2025temporalINR}, we encode the time to the unit circle
\begin{equation}
    t \mapsto (\cos 2\pi t,\ \sin 2\pi t),
\end{equation}
before constructing the spatio-temporal point $\mathbf{x}$.

The training of the neural distance field consisting of parameters $\theta$, $\phi$ and $\mathbf{z}_r$ is over the joint loss
\begin{equation}
\mathcal{L} = \sum_{n}\sum_{\mathbf{x}}
\left[
\big\lVert
\clamp{f_\theta\left(\mathbf{x}, g_\phi(\mathbf{c}^{(n)}), \mathbf{z}_r^{(n)}\right)}
- \clamp{\SDF(\mathbf{x})}
\big\rVert_1
+ \lambda \lVert \mathbf{z}_r^{(n)} \rVert_2
\right],
\end{equation}
where $[x]_{-\delta}^{\delta} := \min(\delta, \max(-\delta, x))$ is a clamp function, restricting the distance from the true surface where the network learns the SDF values. 
The summations run over all patients $(n)$ and all spatio-temporal points $(\mathbf{x})$ per patient.

\textbf{Structual Causal Model.}
Our causal model is built on a directed acyclic graph (DAG) consisting of six clinical variables (\cref{fig:method}(a)). Age and sex are roots of the graph. BMI, smoking status, systolic blood pressure (SBP) and heart rate (HR) are intermediate variables. The terminal node is the $3\mathrm{D}{+}t$ cardiac anatomy. Each edge is grounded in cardiovascular epidemiology research.\footnote{Higher BMI causally raises SBP~\cite{lyall2017bmi}, and sex exerts differential causal effects on BMI and SBP \cite{censin2019obesity}.
Obesity was also linked with cardiovascular remodelling both independently and cojoined \cite{vonjeinsen2020joint}.
Elevated SBP drives atrial remodelling, observed in the Framingham Study
\cite{vaziri1995influence} and supported by Sun et al. \cite{sun2022hypertension} who causally links hypertension to larger LA volumes.
BMI acts on the atrium both through these mediators and directly, it was the
strongest determinant of LA size in the Framingham Study \cite{vaziri1995influence} and is a risk factor for AF \cite{chatterjee2017genetic}. 
Smoking acts via sympathetic activation, acutely raising HR and blood pressure \cite{dimitriadis2022acute}, and is a causal risk factor for atrial fibrillation (AF) \cite{wu2024smoking}, giving its edges to HR, SBP and the anatomy. HR has shown to be a cardiac remodelling factor \cite{ma2023resting} and to have a causal relation with AF \cite{klevjer2023resting}.}

Methodologically we follow the deep structural causal model (DSCM) framework
\cite{pawlowski2020dscm}, modelling each variable $c_i$ as a function $f_i$ of its parents $c_{pa(i)}$ and exogenous noise $\varepsilon_i$,
\begin{equation}
    c_i = f_i (\varepsilon_i, c_{pa(i)}).
\end{equation}
Crucially, each $f_i$ is conditionally-invertible w.r.t. $\varepsilon_i$. The continuous variables (BMI, HR, SBP, and age) are modelled by conditional neural spline flows \cite{Durkan2019NSF}. This maps a standard-normal $\varepsilon_i$ to the given variable conditioned on its parents, the flow for age is unconditional. These continuous variables are standardized prior to modelling. Smoking uses a conditional Bernoulli mechanism and the root female is drawn from its empirical marginal. 

As the graph is a fully observed DAG, the joint likelihood factorises over nodes, therefore each mechanism can be fitted independently by maximum likelihood,
\begin{equation}
    \min_{f_i}\ -\sum_n \log p\!\left(c_i^{(n)} \mid c_{\mathrm{pa}(i)}^{(n)}\right).
\end{equation} 
The SCM is fitted separate to the neural distance field and is the basis of the causal counterfactual generation.

\textbf{Counterfactual generation.}
Given a test subject with observed cardiac anatomy $A_{obs}$ and observed clinical values $\mathbf{c}_{obs}$, the counterfactual anatomy is generated under the intervention $\mathrm{do}(c_i = a)$ in three steps.

\textit{Latent recovery.} With the networks $f_\theta$, $g_\phi$ and the SCM frozen, we recover the latent vector for a test subject by minimizing the reconstruction objective
\begin{equation}
\mathbf{z}_r^{*} = \argmin_{\mathbf{z}_r}
\sum_{\mathbf{x}}
\big\lVert
\clamp{f_\theta(\mathbf{x}, g_\phi(\mathbf{c}), \mathbf{z}_r)}
- \clamp{\SDF(\mathbf{x})}
\big\rVert_1
+ \lambda \lVert \mathbf{z}_r \rVert_2,
\end{equation}
optimizing for $\mathbf{z}_r^*$ over all spatio-temporal points for the test subject. This recovers the subject specific latent code, representing their cardiac anatomy.

\textit{Causal propagation.} The intervention is applied within the SCM by abduction–action–prediction. The exogenous noise variables are abducted from $\mathbf{c}_{\mathrm{obs}}$ using the inverse function of each node. The mechanism of the intervened variable is substituted by its $\mathrm{do}$-value. This value is propagated forward through the graph in topological order, recomputing each descendant from its parent nodes, reusing its abducted exogenous variable. This yields a counterfactual clinical vector $\mathbf{c}_{\mathrm{cf}}$ in which variables downstream of the intervention shift consistently.

\textit{Counterfactual decoding.} The counterfactual clinical vector is re-encoded, $\hat{\mathbf{z}}_c = g_\phi(\mathbf{c}_{\mathrm{cf}})$, and the counterfactual anatomy is decoded with the shape latent and counterfactual clinical laten held fixed,
\begin{equation}
    A_{\mathrm{cf}} = \left\{\, \mathbf{x} :\ f_\theta\!\left(\mathbf{x};\, \hat{\mathbf{z}}_c,\, \mathbf{z}_r^*\right) = 0 \,\right\}.
\end{equation}
Fixing $\mathbf{z}_r^*$ preserves subject identity, making the generated sequence a counterfactual for that individual rather than a population sample.

\textbf{Implementation details and data processing.}
The networks $f_\theta$ and $g_\phi$ are implemented as multi-layer perceptrons, where $f_\theta$ follows the architecture of DeepSDF \cite{park2019deepSDF} and $g_\phi$ is a small network with two hidden layers of 128 neurons. The latent vectors have dimensions 64 and 128 for $\mathbf{z}_r$ and $\mathbf{z}_c$, respectively.

The pre-processing of the surfaces follows \cite{Sorensen2022nudf}, where surfaces for all patients are aligned within the unit sphere using the scan at $t=0\%$ by rigid transformations.
Points for training are randomly sampled by sampling 10 000 points within the unit sphere and 100 000 points in the vicinity of the surface for each time step, accompanied by the signed point-to-surface distance.
All surfaces are constructed by evaluating the network on a uniform grid of size $128^3$ per time-point. A surface can then be extracted using marching cubes \cite{lorensen1987marching}.
\section{Results}
\textbf{Dataset.}
We evaluate our method over LA complexes from 960 4D CT-scans from \textit{The Copenhagen General Population Study}, taken at Rigshospitalet, Copenhagen, Denmark. Each scan contains 20 volumetric frames of the heart equally spaced over a cardiac cycle ($t {=} \{0\%, 5\%, \dots, 95\%\}$). The LA, LAA and PVs surfaces are extracted on each frame using an in-house segmentation model trained on data from ${\sim}2000$ late diastole cardiac CT images from patients not included in this study \cite{loft2026la_population}.
The data is split randomly with $679/85/196$ patients for training, validation and testing, respectively.

We compare our method, denoted by SCM, against the baseline spatio-temporal neural distance field, where the clinical variables does not pass through an SCM before encoding. The two methods have the same clinical input variables, that are normalized prior to encoding. 


\begin{figure}[htbp!]
    \centering
    \includegraphics[width=\linewidth]{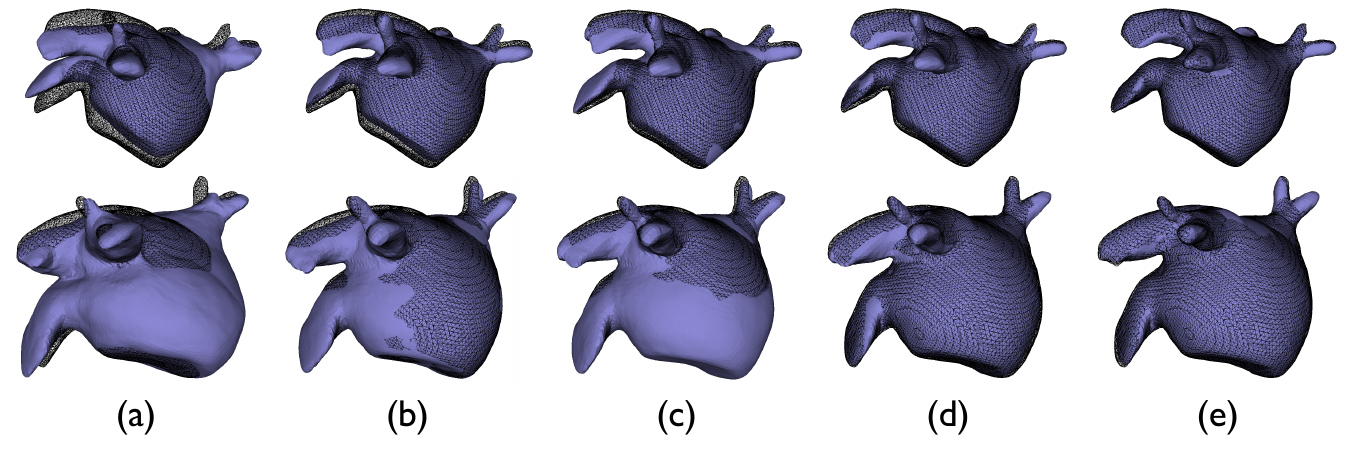}
    \caption{Counterfactual generation for changing age for a single patient. The original surface is represented with the black mesh and the counterfactuals with coloured surfaces. The original has $\mathrm{age} = 83$, while the counterfactuals are generated with interventions $\mathrm{do}(\mathrm{age}{=}\alpha)$, where $\alpha \in \{40,50,60,70,80\}$ for (a) - (e), respectively. The top row is at $t=0\%$ (contraction phase) and the bottom row is at $t=35\%$ (dilation phase).}
    \label{fig:cf_age}
\end{figure}

\begin{table}[htbp!]
\centering
\setlength{\tabcolsep}{6pt}
\small
\caption{Dose response comparisons for the interventions $\mathrm{do}(\mathrm{age}{=}\alpha)$, where $\alpha \in \{40,50,60,70,80\}$. Mean left-atrial volume at $t=0\%$ (Vol) and fraction change (FC) comparing the Baseline model, the SCM model, and the test data whose age is within $\pm5$\,y of the intervened age (GT$\pm$5y).}
\label{tab:age_dose}
\begin{tabular}{@{}ll ccccc @{}}
\toprule
\multicolumn{2}{@{}l}{Age (y)} & 40 & 50 & 60 & 70 & 80 \\
\midrule
\multirow{3}{*}{Vol (ml)}
  & GT$\pm$5y & 81.8 & 79.6 & 82.6 & 84.0 & 109.7 \\
  & Baseline  & 74.3 & 75.2 & 76.7 & 79.2 & 81.4  \\
  & SCM       & \textbf{76.6} & \textbf{78.1} & \textbf{78.6} & \textbf{83.8} & \textbf{91.8} \\
\addlinespace
\multirow{3}{*}{FC (\%)}
  & GT$\pm$5y & 40.1 & 39.1 & 37.0 & 36.2 & 28.3 \\
  & Baseline  & \textbf{39.0} & 37.9 & 36.3 & 34.5 & 32.8 \\
  & SCM       & 44.3 & \textbf{39.0} & \textbf{37.3} & \textbf{34.7} & \textbf{32.5} \\
\bottomrule
\end{tabular}
\end{table}

\begin{figure}[htbp!]
    \centering
    \includegraphics[width=\linewidth]{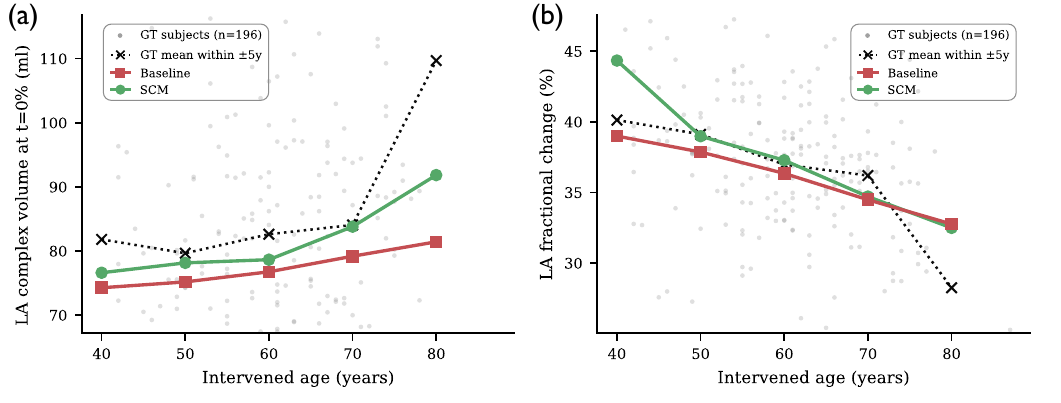}
    \caption{Dose response comparisons for the interventions $\mathrm{do}(\mathrm{age}{=}\alpha)$, where $\alpha \in \{40,50,60,70,80\}$. The crosses indicates the mean of the population within 5 years of the given age. (a) mean of volumes at $t=0\%$, (b) fractional change of volume over cardiac cycle. It should be noted that there are few subjects in limit age groups, this makes the GT mean unreliable.}
    \label{fig:age_dose_response}
\end{figure}

\begin{table}[htbp!]
  \centering
  \setlength{\tabcolsep}{6pt}

  \caption{LA-complex volume distributions at end-diastole (ml) for the two
  ground-truth subgroups and the two counterfactual models under   $\mathrm{do}(\mathrm{female}{=}1)$.  Counterfactuals are computed on the $n{=}85$ GT-male test subjects. Bold indicates closest to GT female distribution between the baseline and SCM models.}
  \label{tab:female_cf_distributions}
  \begin{tabular}{lccccc}
  \toprule
  Group & $n$ & Mean & Std & Median & Range \\
  \midrule
  GT male                            &  85 & 94.76 & 20.78 & 91.48 & [54.25, 160.90] \\
  GT female                          & 111 & 76.03 & 17.66 & 72.41 & [45.63, 169.36] \\
  Baseline  &  85 & 72.91 & \textbf{17.37} & 69.34 & [35.65, 128.19] \\
  SCM       &  85 & \textbf{76.87} & 20.83 & \textbf{71.39} & \textbf{[42.15, 164.20]} \\
  \bottomrule
  \end{tabular}
  \end{table}
  
\begin{figure}[htbp]
    \centering
    \includegraphics[width=\linewidth]{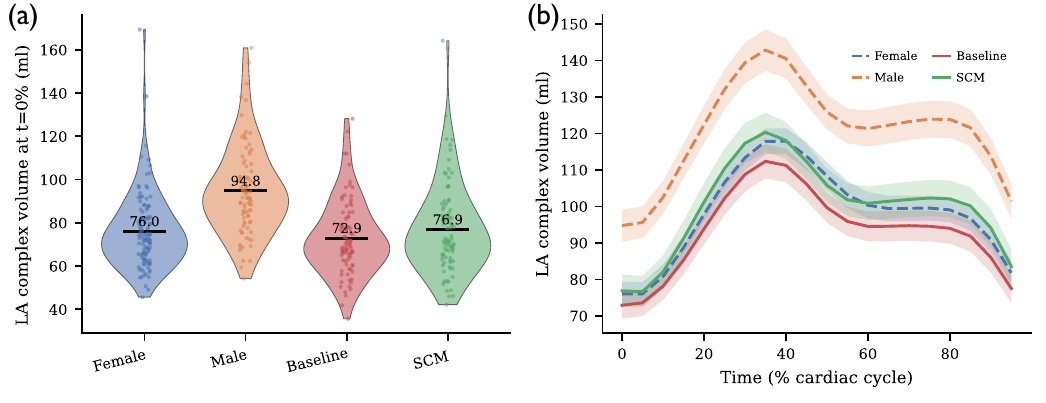}
    \caption{Volume distribution comparisons for counterfactuals with $\mathrm{do}(\mathrm{female}=1)$ on the male test set. (a) volume distribution of the LA complex at $t=0\%$, the black line indicates the mean of the distribution. (b) mean volume over time, shaded area indicates 95\% CI for the group mean.}
    \label{fig:do_female}
\end{figure}

\textbf{Counterfactual generation.}
A representative example of counterfactual generation is shown in \cref{fig:cf_age} for a subject aged 83. Higher-age interventions yield anatomies close to the factual anatomy, whereas lower-age interventions produce anatomies that differs more from the factual anatomy. Generally, at $t=0\%$, the LA body shrinks, the LAA remains the same size but moves with the LA and the PVs enlarge. A video showing the movement over a cardiac cycle for all five ages is found as \textit{Supplementary Video 1} (also available on our GitHub). 
The example corresponds well with the general trend for intervening on subject age shown in \cref{fig:age_dose_response} and in \cref{tab:age_dose}, 
where the volume at $t=0\%$ increases with higher age both for the models and for the factual test data. Higher volume at $t=0\%$ corresponds to a worse contraction of the LA complex. The LA fractional change is decreasing with age, which can be attributed to smaller volume differences between $t=0\%$ and $t=35\%$ with increasing age.

The intervention $\mathrm{do}(\mathrm{female}=1)$ on the male test set is visualized on distribution level in \cref{fig:do_female} and with descriptive statistics in \cref{tab:female_cf_distributions}. Both for the baseline model and for our model the volume distribution is shifted down closer to the female distribution. 
As a consistency check, we test whether original anatomies can be recovered from counterfactual anatomies by doing the inverse intervention. Starting from counterfactual female anatomies we re-fit $\mathbf{z}_r^*$ from the input and then intervene with $\mathrm{do}(\mathrm{female}=0)$ to return back to male. This is trivial for the baseline as it collapses to encode-decode.
The recovered surfaces from our model closely match the originals at
$t=0\%$, with volume difference $\Delta V=-0.219$ ml ($\mathrm{std}=2.169$ ml) and Chamfer distance $CD = 0.13$ mm ($\mathrm{std}=0.276$ mm) -- below the voxel spacing of the CT-scans surfaces originate from.

\section{Discussion and Conclusion}

In the generated age counterfactuals, intervening with a younger age yields a smaller atrium during  the contraction phase ($t=0\%$), as well as a larger LA fractional change. This reproduces a known age-related decline in atrial function~\cite{Vasconcellos2021LAref}. 
The subject-specific morphology, appendage shape and pulmonary-vein count and configuration, is preserved across interventions, as the subject latent code $\mathbf{z}_r$ is fixed.
This supports the intuition that the two latent codes encode complementary information.
At the distribution level the SCM counterfactuals resemble the real data distribution more closely than the baseline model. For the age intervention the SCM model show a larger response to the changing age when the changes propagate down through the graph before the condition encoding. For the female intervention the baseline overshoots the effect, while the distribution from the SCM model closely resembles the actual female distribution keeping the range close to the actual distribution range. 
The ability to do inverse interventions further supports that the soundness of the SCM counterfactuals, as it recovers male anatomy from female counterfactuals with volumes only marginally different from originals. This approximate reversibility indicates that changing the clinical encoding does not alter the structure of the LA complex.

While we only include counterfactual generation from parent nodes, the SCM also supports causal generation on intermediate or leaf nodes. Interventions on leaf nodes will not pass through the SCM and thus collapse to the baseline, whereas interventions on intermediate nodes will pass through the SCM. We focus on age and sex because the data exhibit a clear trend that the distributions of counterfactuals can be compared against.

Several limitations for our study should be noted. The DAG our SCM is build upon has not been established as a causal graph for cardiac anatomy before. We justify each edge from cardiovascular research, however not all are firmly causal and some edges could not be included due to the inherent limitations of a DAG, e.g. heart rate and blood pressure are linked, but the edge orientation is unclear. 
Furthermore, evaluation of counterfactuals is inherently challenging as no ground truth exists for any intervention. Longitudinal scans could provide a comparison for interventions both for the SCM and generated anatomy, though these are not available for this data.
Looking ahead, extending the framework to larger, disease-spanning could provide a controllable source of causal data augmentation for under-represented anatomies, supporting fairer and more robust downstream cardiac models.
 
In conclusion, we present a causal generation framework for 4D cardiac anatomy. This is, to our knowledge, the first counterfactual generation of 4D cardiac anatomy, and the first integration of a structural causal model with an implicit neural representation as its image mechanism. The resulting model produces counterfactual anatomies that follow population trends, are approximately reversible under inverse intervention, and preserve subject identity, while remaining causally consistent.

\subsubsection{Disclosure of interests.} The authors have no competing interests to declare that are relevant to the content of this article.

%
%
%
%
\bibliographystyle{splncs04}
\bibliography{references.bib}

\end{document}